\documentclass{article}

\usepackage{times}
\usepackage{graphicx} 
\usepackage{subfigure} 

\usepackage{natbib}

\usepackage{algorithm}
\usepackage{algorithmic}

\usepackage[accepted]{icml2017}

\usepackage{url}
\usepackage{xcolor}
\usepackage{paralist}
\usepackage{amssymb}
\usepackage{amsmath}
\usepackage{wrapfig}

\usepackage[abs]{overpic}
\usepackage{calc}
\newcommand{\real}{\mathbb{R}}
\newcommand{\myvec}[1]{\mathbf{#1}}

\newcommand{\vh}{\myvec{h}}

\newcommand{\vx}{\myvec{x}}

\newcommand{\E}{\mathbb{E}}

\newcommand{\calX}{{\cal X}}

\newcommand{\be}{\begin{equation}}
\newcommand{\ee}{\end{equation}}
\newcommand{\bea}{\begin{eqnarray}}
\newcommand{\eea}{\end{eqnarray}}
\newcommand{\beaa}{\begin{eqnarray*}}
\newcommand{\eeaa}{\end{eqnarray*}}

\DeclareMathAlphabet{\mathpzc}{OT1}{pzc}{m}{n}

\DeclareMathOperator*{\argmin}{arg\,min}

\newcommand{\xstar}{\myvec{x}^\star}
\newcommand{\given}{\mid}

\icmltitlerunning{Learning to Learn without Gradient Descent by Gradient Descent}

\begin{document} 

\twocolumn[
\icmltitle{
\begin{minipage}{\widthof{Learning to Learn without Gradient Descent}}
\begin{flushright}
Learning to Learn without Gradient Descent\\
by Gradient Descent
\end{flushright}
\end{minipage}
}



\begin{icmlauthorlist}
\icmlauthor{Yutian Chen}{deepmind}
\icmlauthor{Matthew W.~Hoffman}{deepmind}
\icmlauthor{Sergio G\'omez Colmenarejo}{deepmind}
\icmlauthor{Misha Denil}{deepmind}
\icmlauthor{Timothy P.~Lillicrap}{deepmind}
\icmlauthor{Matt Botvinick}{deepmind}
\icmlauthor{Nando de Freitas}{deepmind} 
\end{icmlauthorlist}

\icmlaffiliation{deepmind}{DeepMind, London, United Kingdom}

\icmlcorrespondingauthor{Yutian Chen}{yutianc@google.com}

\icmlkeywords{learning to learn, learning to optimize, meta learning,
              black box optimization, global optimization}

\vskip 0.3in
]



\printAffiliationsAndNotice{}  

\begin{abstract} 



We learn recurrent neural network optimizers trained on simple synthetic functions by gradient descent. We show that these learned optimizers exhibit a remarkable degree of transfer in that they can be used to
efficiently optimize a broad range of derivative-free black-box functions, including Gaussian process bandits, simple control objectives, global optimization benchmarks and hyper-parameter tuning tasks.  
Up to the training horizon, the learned optimizers learn to trade-off exploration and exploitation, and compare favourably with heavily engineered Bayesian optimization packages for hyper-parameter tuning.

\end{abstract}

\section{Introduction}

Findings in developmental psychology have revealed that infants are endowed with a small number of separable systems of core knowledge for reasoning about objects, actions, number, space, and possibly social interactions \citep{spelke2007core}. These systems enable infants to learn many skills and acquire knowledge rapidly. The most coherent explanation of this phenomenon is that the learning (or optimization) process of evolution has led to the emergence of components that enable fast and varied forms of learning. In psychology, \emph{learning to learn} has a long history  \citep{ward1937reminiscence,harlow1949formation,kehoe1988layered}. 

Inspired by this, many researchers have attempted to build agents capable of learning to learn \citep{schmidhuber:1987,naik:1992,thrun:1998,hochreiter:2001,santoro:2016,Duan2016,Wang2016learning,Ravi2017optimization,Li2017learning}. The scope of research under the umbrella of learning to learn is very broad. The learner can implement and be trained by many different algorithms, including gradient descent, evolutionary strategies, simulated annealing, and reinforcement learning.  

For instance, one can learn to learn by gradient descent by gradient descent, or learn local Hebbian updates by gradient descent \citep{andrychowicz2016learning,Bengio+al-92}. In the former, one uses supervised learning at the meta-level to learn an algorithm for supervised learning, while in the latter, one uses supervised learning at the meta-level to learn an algorithm for unsupervised learning. 

Learning to learn can be used to learn both models and algorithms. In \citet{andrychowicz2016learning} the output of meta-learning is a trained recurrent neural network (RNN), which is subsequently used as an optimization algorithm to fit other models to data.  In contrast, in  \citet{ZophLe2017} the output of meta-learning can also be an RNN model, but this new RNN is subsequently used as a model that is fit to data using a classical optimizer. In both cases the output of meta-learning is an RNN, but this RNN is interpreted and applied as a model or as an algorithm. In this sense, learning to learn with neural networks blurs the classical distinction between models and algorithms. 

In this work, the goal of meta-learning is to produce an algorithm for global black-box optimization. 
Specifically, we address the problem of finding a global minimizer of an unknown (black-box) loss function~$f$. That is, we wish to compute $
	\xstar = \argmin_{\vx \in \calX} f(\vx)\,,
$
where~$\calX$ is some search space of interest. The black-box function~$f$ is not available to the learner in simple closed form at test time, but can be evaluated at a query point~$\vx$ in the domain.  This evaluation produces either deterministic or stochastic outputs~${y \in \real}$ such that~$f(\vx) = {\E[y \given f(\vx)]}$. In other words, we can only observe the function~$f$ through unbiased noisy point-wise observations~$y$.

Bayesian optimization is one of the most popular black-box optimization methods \citep{Brochu:2009,Snoek:2012,Outoftheloop}.
It is a sequential model-based decision making approach with two components. 
The first component is a probabilistic model, consisting of a prior distribution that captures our beliefs about the behavior of the unknown objective function and an observation model that describes the data generation mechanism. The model can be a Beta-Bernoulli bandit, a random forest, a Bayesian neural network, or a Gaussian process (GP) \citep{Outoftheloop}. Bayesian optimization is however often associated with GPs, to the point of sometimes being referred to as GP bandits \citep{Srinivas:2010}.
 
The second component is an acquisition function, which is optimized at each step so as to trade-off exploration and exploitation. Here again we encounter a huge variety of strategies, including Thompson sampling,  information gain, probability of improvement, expected improvement, upper confidence bounds \citep{Outoftheloop}. 
The requirement for optimizing the acquisition function at each step can be a significant cost, as shown in the empirical section of this paper. It also raises some theoretical concerns \citep{Wang:2014aistats}.

In this paper, we present a learning to learn approach for global optimization of black-box functions and contrast it with Bayesian optimization. In the meta-learning phase, we use a large number of differentiable functions generated with a GP 
to train RNN optimizers by gradient descent. We consider two types of RNN: long-short-term memory networks (LSTMs) by \citet{hochreiter:1997} and differentiable neural computers (DNCs) by \citet{dnc2016}. 

During meta-learning, we choose the horizon (number of steps) of the optimization process. We are therefore considering the finite horizon setting that is popular in AB tests \cite{Kohavi:2009,Scott:2010} and is often studied under the umbrella of best arm identification in the bandits literature~\citep{Bubeck:2009,Gabillon:2012}.

The RNN optimizer learns to use its memory to store information about previous queries and function evaluations, and learns to access its memory to make decisions about which parts of the domain to explore or exploit next. That is,
by unrolling the RNN, we generate new candidates for the search process. The experiments will show that this process is much faster than applying standard Bayesian optimization, and in particular it does not involve either matrix inversion or optimization of acquisition functions. 

In the experiments we also investigate distillation of acquisition functions to guide the process of training the RNN optimizers, and the use of parallel optimization schemes for expensive training of deep networks.

The experiments show that the learned optimizers can transfer to optimize a large and diverse set of black-box functions arising in global optimization, control, and hyper-parameter tuning. Moreover, withing the training horizon, the RNN optimizers are competitive with state-of-the-art heavily engineered packages such as Spearmint, SMAC and TPE \cite{snoek-warping-2014,Hutter:smac,bergstra2011algorithms}

\section{Learning Black-box Optimization}
\label{sec:algorithm}

\begin{figure*}[tbph]
    \centering
    \includegraphics[width=0.7\textwidth]{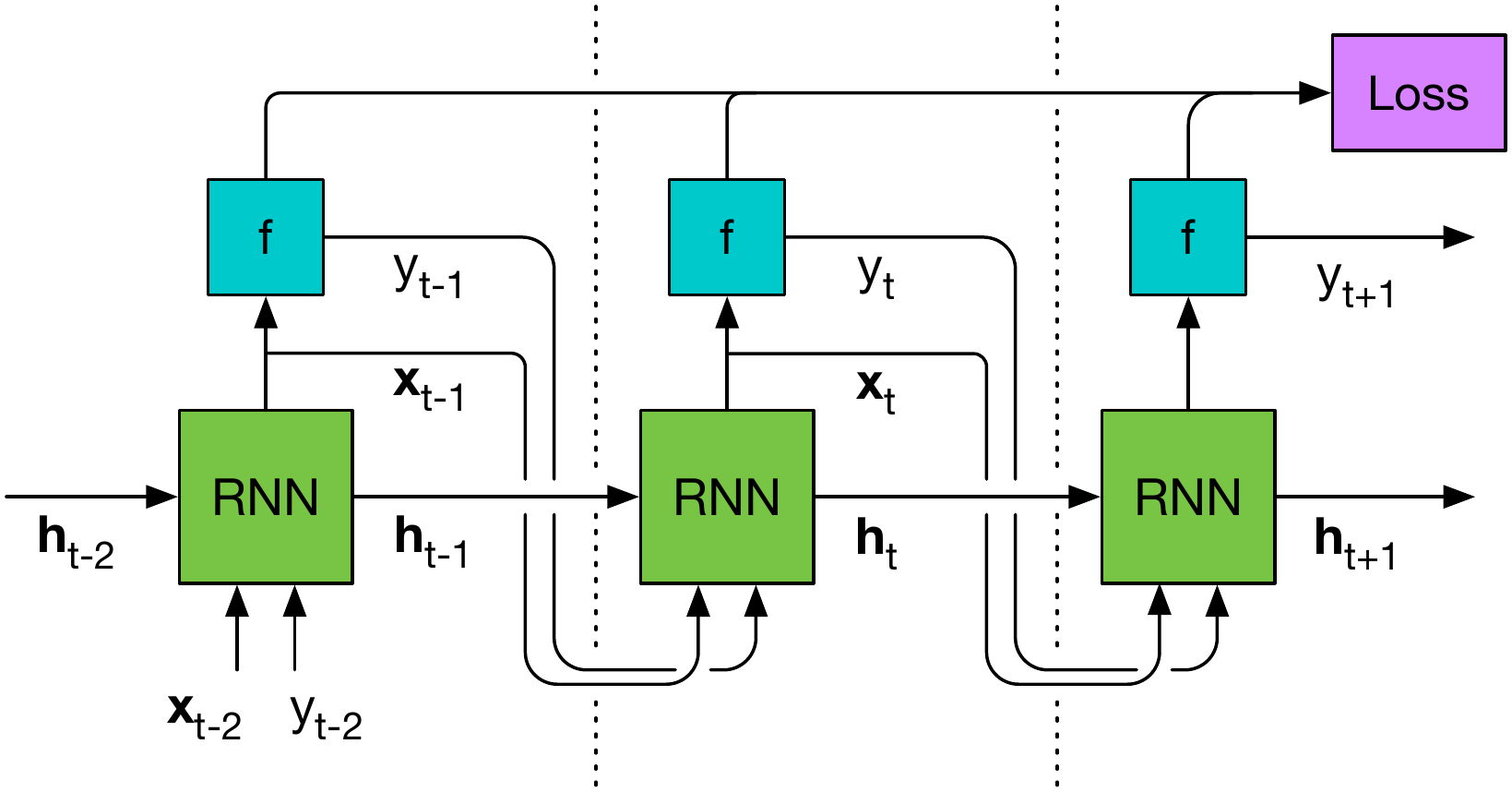}
    \caption{Computational graph of the learned black-box optimizer unrolled over multiple steps. The learning process will consist of differentiating the given loss with respect to the RNN parameters }
    \label{fig:graph}
\end{figure*}

A black-box optimization algorithm can be summarized by the following loop:
\begin{compactenum}
    \item Given the current state of knowledge $\vh_t$ propose a query point $\vx_t$
    \item Observe the response $y_t$
    \item Update any internal statistics to produce $\vh_{t+1}$
\end{compactenum}
This easily maps onto the classical frameworks presented in the previous section where the update step computes statistics and the query step uses these statistics for exploration. In this work we take this framework as a starting point and define a combined update and query rule using a recurrent neural network parameterized by $\theta$ such that
\begin{align}
    \vh_t, \vx_t &= \mathrm{RNN}_\theta(\vh_{t-1}, \vx_{t-1}, y_{t-1}),\\
    y_t &\sim p(y\given \vx_t) \enspace.
\end{align}
Intuitively this rule can be seen to update its hidden state using data from the previous time step and then propose a new query point. In what follows we will apply this RNN, with shared parameters, to many steps of a black-box optimization process. An example of this computation is shown in Figure~\ref{fig:graph}. Additionally, note that in order to generate the first query $\vx_1$ we arbitrarily set the initial ``observations'' to dummy values $\vx_0=\mathbf{0}$ and $y_0=0$; this is a point we will return to in Section~\ref{sec:parallel}. 


\subsection{Loss Function}

Given this rule we now need a way to learn the parameters $\theta$ with stochastic gradient descent for any given distribution of differentiable functions $p(f)$. Perhaps the simplest loss function one could use is the loss of the final iteration: $L_\text{final}(\theta)=\E_{f,y_{1:T-1}}[f(\vx_T)]$ for some time-horizon $T$. This loss was considered by \citet{andrychowicz2016learning} in the context of learning first-order optimizers, but ultimately rejected in favor of the summed loss
\begin{align}
    L_\text{sum}(\theta)
    &= \E_{f,y_{1:T-1}}\left[\sum_{t=1}^T f(\vx_t) \right] .
\end{align}
A key reason to prefer $L_\text{sum}$ is that the amount of information conveyed by $L_\text{final}$ is temporally very sparse. By instead utilizing a sum of losses to train the optimizer we are able to provide information from every step along this trajectory. Although at test time the optimizer typically only has access to the observation $y_t$, at training time the true loss can be used. Note that optimizing the summed loss is equivalent to finding a strategy which minimizes the expected cumulative regret. Finally, while in many optimization tasks the loss associated with the best observation $\min_t f(\vx_t)$ is often desired, the cumulative regret can be seen as a proxy for this quantity.

By using the above objective function we will be encouraged to trade off exploration and exploitation and hence globally optimize the function $f$. This is due to the fact that in expectation, any method that is better able to explore and find small values of $f(\vx)$ will be rewarded for these discoveries. However the actual process of optimizing the above loss can be difficult due to the fact that nothing explicitly encourages the optimizer itself to explore. 

We can encourage exploration in the space of optimizers by encoding an exploratory force directly into the meta learning loss function. Many examples exist in the bandit and Bayesian optimization communities, for example
\begin{align}
    L_\text{EI}(\theta) 
    &= -\E_{f,y_{1:T-1}}\left[\sum_{t=1}^T \mathrm{EI}(\vx_t\given y_{1:t-1})\right]
\end{align}
where $\mathrm{EI}(\cdot)$ is the expected posterior improvement of querying $\vx_t$ given observations up to time $t$. This can encourage exploration by giving an explicit bonus to the optimizer rather than just implicitly doing so by means of function evaluations. Alternatively, it is possible to use the observed improvement (OI)
\begin{align}
    L_\text{OI}(\theta)
    &= \E_{f,y_{1:T-1}}\left[\sum_{t=1}^T \min\left\{f(\vx_t) - \min_{i<t}(f(\vx_i)), 0\right\}\right]
\end{align}


We also studied a loss based on GP-UCB \citep{Srinivas:2010} but in preliminary experiments this did not perform as well as the EI loss and is thus not included in the later experiments.

The illustration of Figure~\ref{fig:graph} shows the optimizer unrolled over many steps, ultimately culminating in the loss function. To train the optimizer we will simply take derivatives of the loss with respect to the RNN parameters $\theta$ and perform stochastic gradient descent (SGD). In order to evaluate these derivatives we assume that derivatives of $f$ can be computed with respect to its inputs. This assumption is made only in order to backpropagate errors from the loss to the parameters, but crucially is \emph{not needed at test time}. If the derivatives of $f$ are also not available at training time then it would be necessary to approximate these derivatives via an algorithm such as REINFORCE \citep{williams1992simple}.

\subsection{Training Function Distribution}

To this point we have made no assumptions about the distribution of training functions $p(f)$. In this work we are interested in learning general-purpose black-box optimizers, and we desire our distribution to be quite broad. 

As a result we propose the use of GPs as a suitable training distribution. Under the GP prior, the joint distribution of function values at any finite set of query points follows a multivariate Gaussian distribution \citep{Rasmussen:2006}, and we generate a realization of the training function incrementally at the query points using the chain rule with a total time complexity of $O(T^3)$ for every function sample.

The use of functions sampled from a GP prior also provides functions whose gradients can be easily evaluated at training time as noted above. Further, the posterior expected improvement used within $L_\mathrm{EI}$ can be easily computed~\cite{Mockus:1982} and differentiated as well. Search strategies based on GP losses, such as $L_\mathrm{EI}$, can be thought of as a distilled strategies.
The major downside of search strategies which are based on GP inference is their cubic complexity.

While training with a GP prior grants us the convenience to assess the efficacy of our training algorithm by comparing head-to-head with GP-based methods, it is worth noting that our model can be trained with any distribution that permits efficient sampling and function differentiation. The flexibility could become useful when considering problems with specific prior knowledge and/or side information.

\subsection{Parallel Function Evaluation}
\label{sec:parallel}

\begin{figure}
    \centering
    \includegraphics[width=0.46\textwidth]{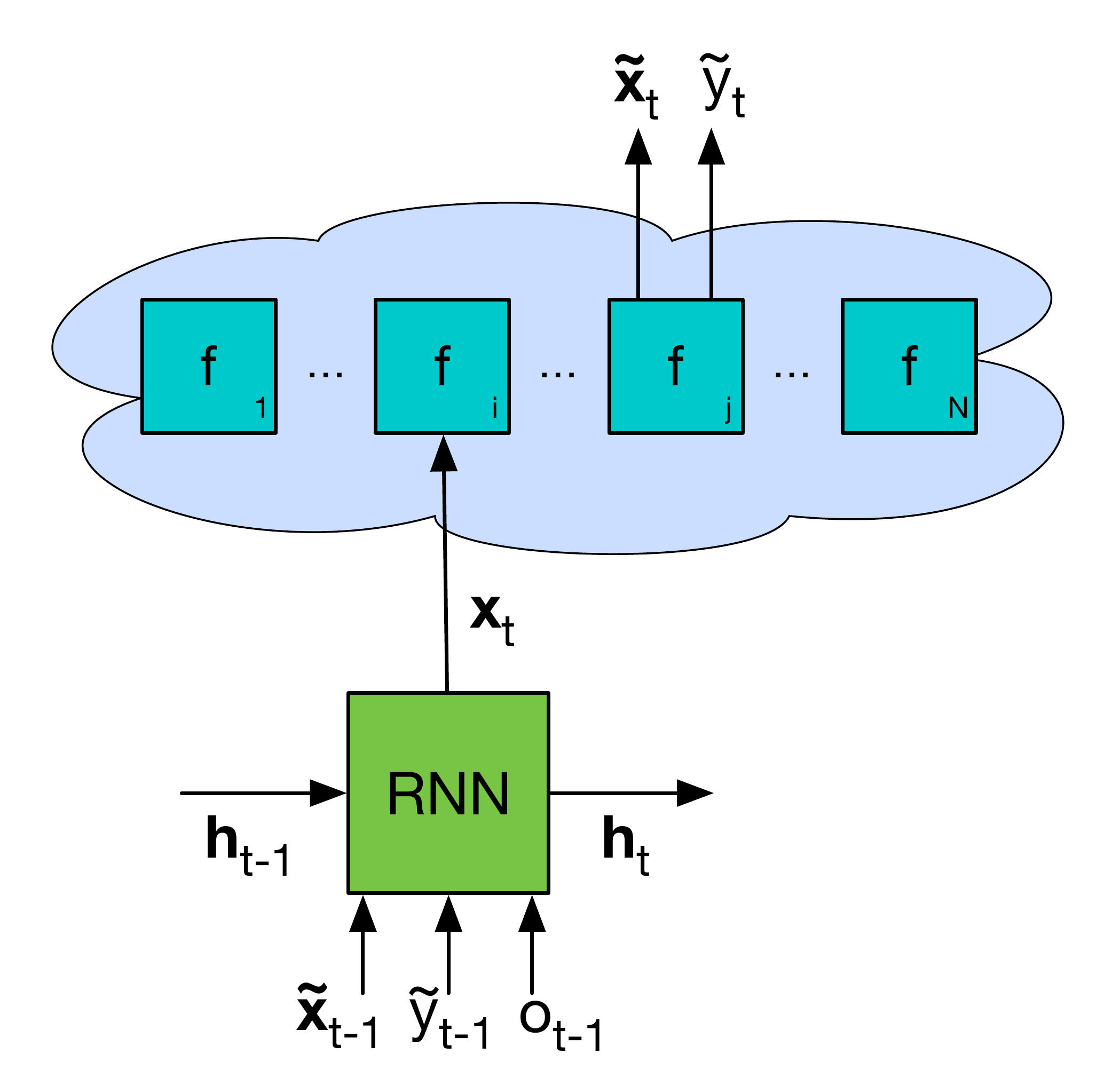}
    \caption{Graph depicting a single iteration of the parallel algorithm with $N$ workers. Here we explicitly illustrate the fact that the query $\vx_t$ is being assigned to the $i$-th worker for evaluation, and that the 
    the next observation pair $(\tilde\vx_t,\tilde y_t)$ is the output of the $j$-th worker, which has completed its function evaluation.}
    \label{fig:parallel}
\end{figure}

The use of parallel function evaluation is a common technique in Bayesian optimization, often used for costly, but easy to simulate functions. For example, as illustrated in the experiments, when searching for hyper-parameters of deep networks, it is convenient to train several deep networks in parallel.

Suppose we have $N$ workers, and that the process of proposing candidates for function evaluation is much faster than evaluating the functions. We augment our RNN optimizer's input with a binary variable $o_t$ as follows:
\begin{align}
    \vh_t, \vx_t &= \mathrm{RNN}_\theta(\vh_{t-1}, o_{t-1}, \tilde{\vx}_{t-1}, \tilde{y}_{t-1}).
\end{align}
For the first $t \leq N$ steps, we set $o_{t-1}=0$, arbitrarily set the inputs to dummy values $\tilde{\vx}_{t-1}=\mathbf{0}$ and $\tilde{y}_{t-1}=0$, and  
generate $N$ parallel queries $\vx_{1:N}$.  
As soon as a worker finishes evaluating a query, the query and its evaluation are  
fed back to the network by setting $o_{t-1}=1$, resulting in a new query $\vx_t$. Figure~\ref{fig:parallel} displays a single iteration of this algorithm.

A well-trained optimizer must learn to condition on $o_{t-1}$ in order to either generate initial queries or generate queries based on past observations.
Another key point to consider is that the batch nature of the optimizer can result in the ordering of queries being permuted: i.e.\ although $\vx_t$ is proposed before $\vx_{t+1}$ it is entirely plausible that $\vx_{t+1}$ is evaluated first. In order to account for this at \emph{training time} and not allow the optimizer to rely on a specific ordering, we simulate a runtime $\Delta_t \sim \mathrm{Uniform}(1 - \sigma, 1 + \sigma)$ associated with the $t$-th query. Observations are then made based on the order in which they complete. 


It is worth noting that the sequential setting is a special case of this parallel policy where $N=1$ and every observation is made with $o_{t-1}=1$. Note also that we have kept the number of workers fixed for simplicity of explanation only. The architecture allows for the number of workers to vary.

It is instructive to contrast this strategy with what is done in parallel Bayesian optimization~\citep{Desautels:2012, Snoek:2012}. There care must be taken to ensure that a diverse set of queries are utilized---in the absence of additional data the standard sequential strategy would propose $N$ queries at the same point. Exactly computing the optimal $N$-step query is typically intractable, and as a result hand-engineered heuristics are employed. Often this involves synthetically reducing the uncertainty associated with outstanding queries in order to simulate later observations. In contrast, our RNN optimizer can store in its hidden state any relevant information about outstanding observations. Decisions about what to store are learned during training and as a result should be more directly related to later losses.



\section{Experiments}

We present several experiments that show the breadth of generalization that is achieved by our learned algorithms.  We train our algorithms to optimize very simple functions---samples from a GP with a fixed length scale---and show that the learned algorithms are able to generalize from these simple objective functions to a wide variety of other test functions that were not seen during training.

\begin{figure*}[t!]%
\centering
\includegraphics[width=\textwidth]{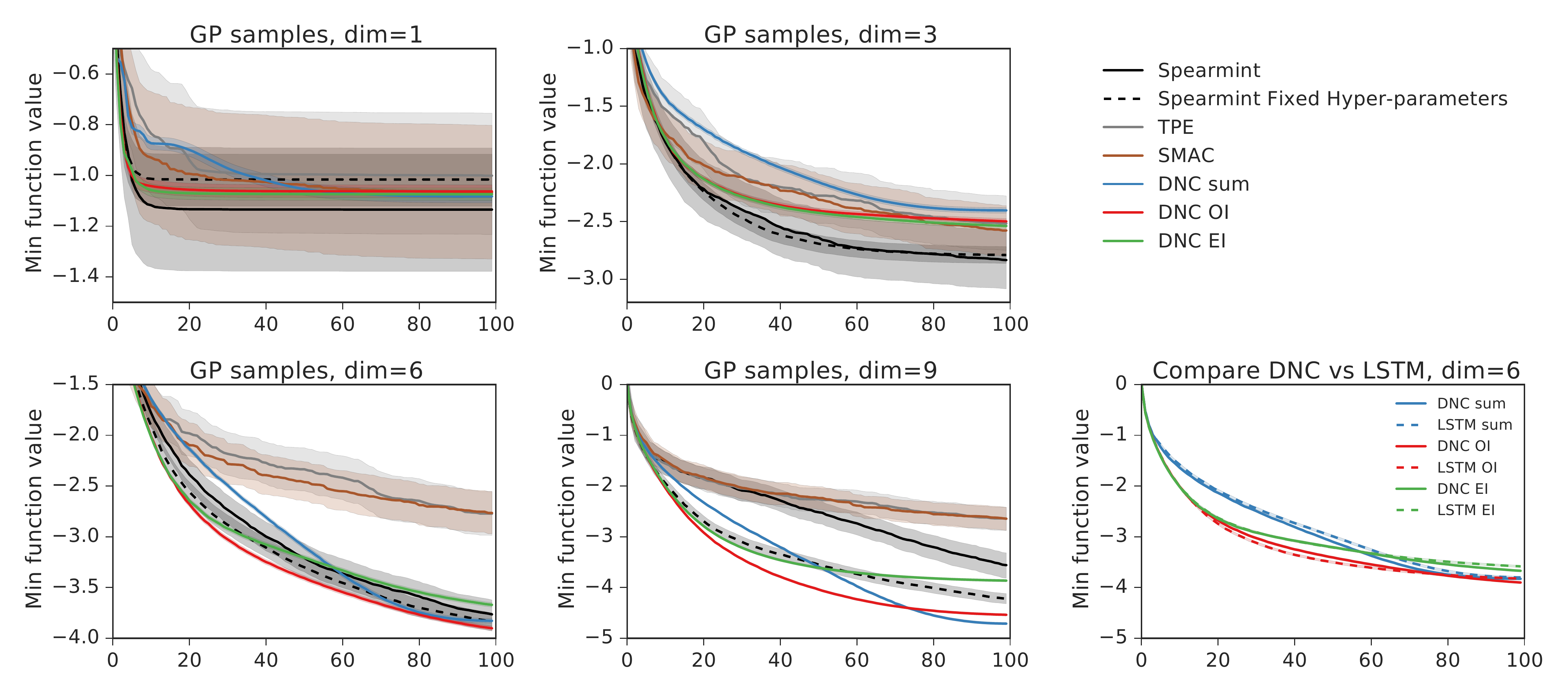}
\caption{Average minimum observed function value, with $95\%$ confidence intervals, as a function of search steps on functions sampled from the training GP distribution. Left four figures: Comparing DNC with different reward functions against Spearmint with fixed and estimated GP hyper-parameters, TPE and SMAC. Right bottom: Comparing different DNCs and LSTMs. As the dimension of the search space increases, the DNC's performance improves relative to the baselines.}
\label{fig:gp}
\end{figure*}

\begin{figure*}%
\centering
\includegraphics[width=\textwidth]{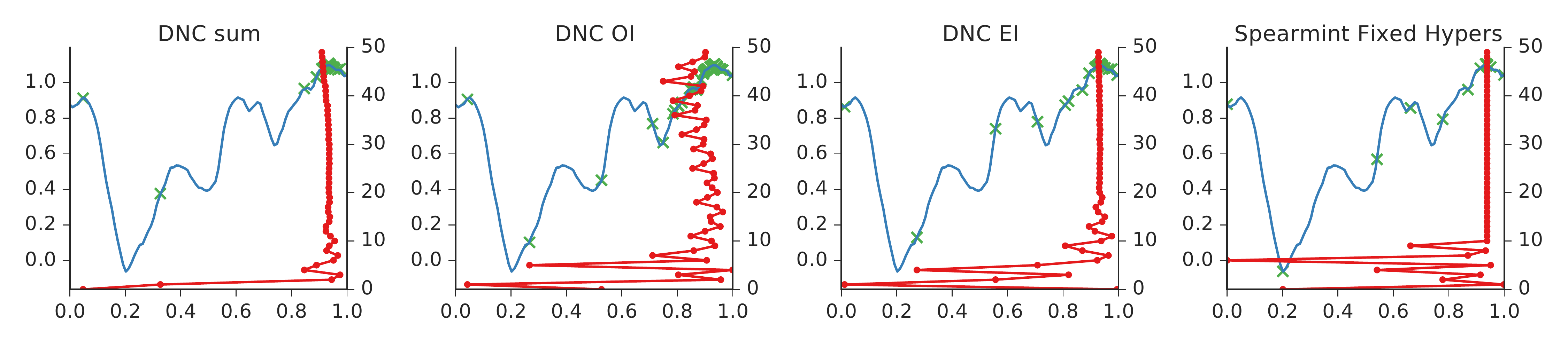}
\caption{How different methods trade-off exploration and exploitation in a one-dimensional example. Blue: Unknown function being optimized. Green crosses: Function values at query points. Red trajectory: Query points over 50 steps.}
\label{fig:trajectories}
\end{figure*}

We experimented with two different RNN architectures: LSTMs and DNCs. However, we found the DNCs to perform slightly (but not significantly) better. For clarity, we only show plots for DNCs in most of the figures.

We train each RNN optimizer with trajectories of $T$ steps, and update the RNN parameters using BPTT with Adam. We use a curriculum to increase the length of trajectories gradually from $T=10$ to $100$. We repeat this process for each of the loss functions discussed in Section~\ref{sec:algorithm}. Hyper-parameters for the RNN optimization algorithm (such as learning rate, number of hidden units, and memory size for the DNC models) are found through grid search during training. When ready to be used as an optimizer, the RNN requires neither tuning of hyper-parameters nor hand-engineering. It is fully automatic.

In the following experiments, \emph{DNC sum} refers to the DNC network trained using the summed loss $L_\mathrm{sum}$, \emph{DNC OI} to the network trained using the loss $L_\mathrm{OI}$, and \emph{DNC EI} to the network trained with the loss $L_\mathrm{EI}$.

We compare our learning to learn approach with popular state-of-the-art Bayesian optimization packages, including Spearmint with automatic inference of the GP hyper-parameters and input warping to deal with non-stationarity \citep{snoek-warping-2014}, Hyperopt (TPE) \citep{bergstra2011algorithms}, and SMAC \citep{hutter2011sequential}. For test functions with integer inputs, we treat them as piece-wise constant functions and round the network output to the closest values. We evaluate the performance at a given search step $t \leq T=100$, according to the minimum observed function value up to step $t$, $\min_{i\leq t} f(\vx_i)$.

\begin{figure*}[t]
\centering
\includegraphics[height=2.95in]{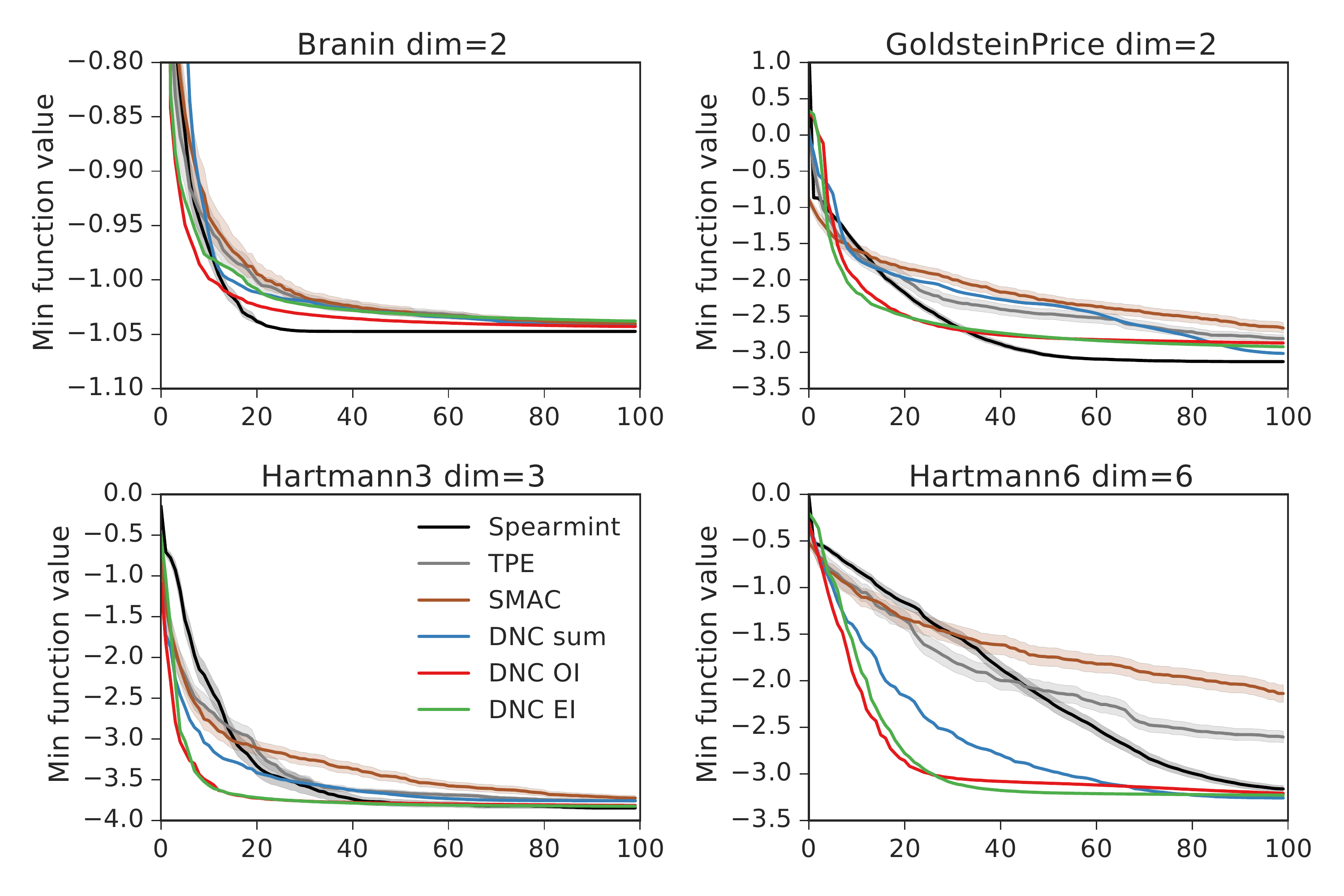}
\rule{1pt}{3in}
\includegraphics[height=2.95in]{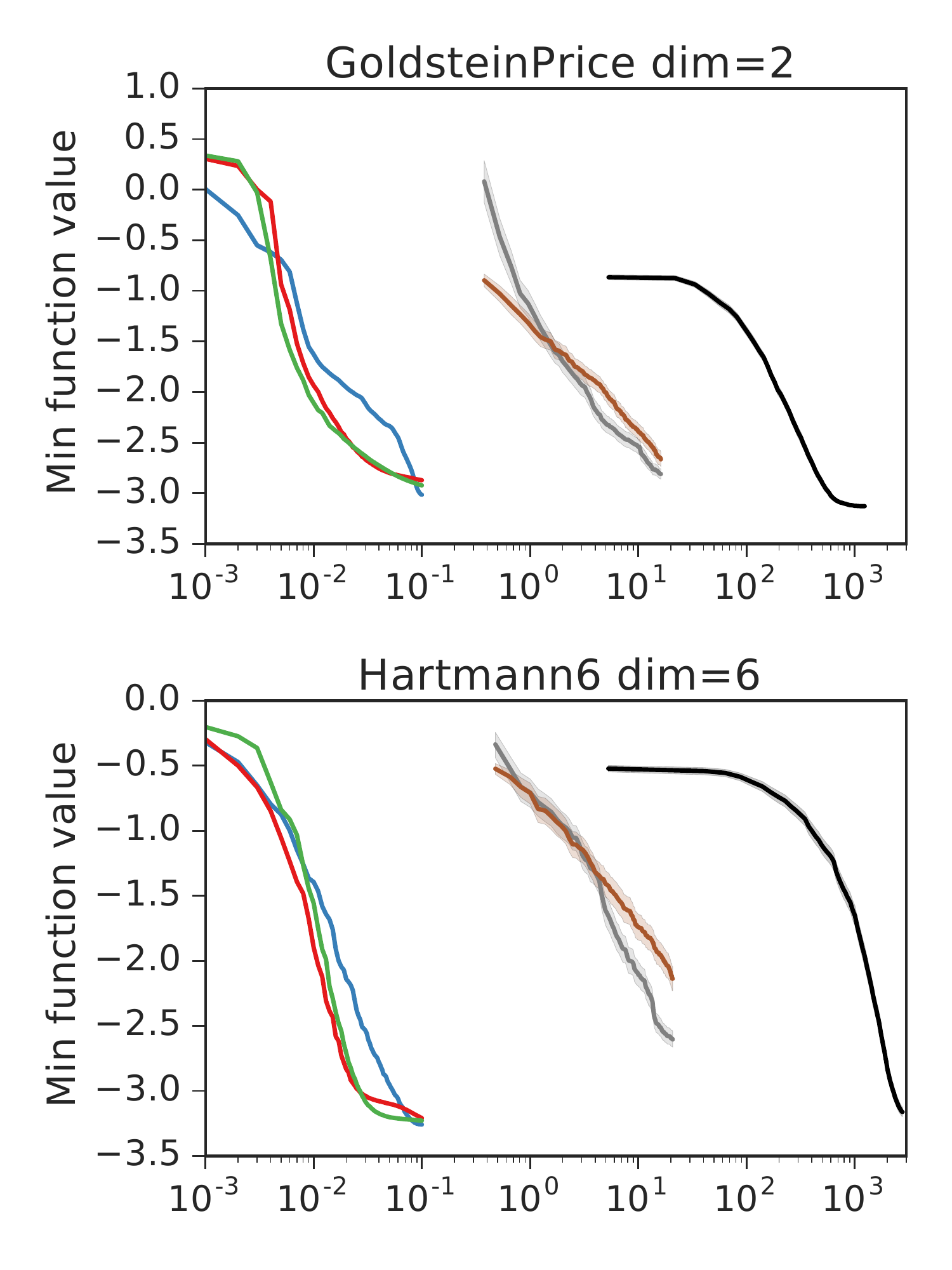}
\caption{Left: Average minimum observed function value, with $95\%$ confidence intervals, as a function of search steps on 4 benchmark functions: Branin, Goldstein price, 3d-Hartmann and 6d-Hartmann. Again we see that as the dimension of the search space increases, the learned DNC optimizers are more effective than the Spearmint, TPE and SMAC packages within the training horizon. 
Right: Average minimum observed function value in terms of the optimizer's run-time (seconds), illustrating the superiority in speed of the DNC optimizers over existing black-box optimization methods.}
\label{fig:benchmarks}
\end{figure*}


\subsection{Performance on Functions Sampled from the Training Distribution}

We first evaluate performance on functions sampled from the training distribution. Notice, however, that these functions are never observed during training. Figure \ref{fig:gp} shows the best observed function values as a function of search step $t$, averaged over $10,000$ sampled functions for RNN models and 100 sampled functions for other models (we can afford to do more for RNNs because they are very fast optimizers). For Spearmint, we consider both the default setting with a prior distribution that estimates the GP hyper-parameters by Monte Carlo and a setting with the same hyper-parameters as those used in training. For the second setting, Spearmint knows the ground truth and thus provides a very competitive baseline. As expected Spearmint with a fixed prior proves to be one of the best models under most settings. When the input dimension is 6 or higher, however, neural network models start to outperform Spearmint. We suspect it is because in higher dimensional spaces, the RNN optimizer learns to be more exploitative given the fixed number of iterations. Among all RNNs, those trained with expected/observed improvement perform better than those trained with direct function observations.

Figure \ref{fig:trajectories} shows the query trajectories $\vx_t$, $t=1,\ldots,100$, for different black-box optimizers in a one-dimensional example. All of the optimizers explore initially, and later settle in one mode and search more locally. The DNCs trained with EI behave most similarly to Spearmint. DNC with direct function observations (DNC sum) tends to explore less than the other optimizers and often misses the global optimum, while the DNCs  trained with the observed improvement (OI) keep exploring even in later stages.


\subsection{Transfer to Global Optimization Benchmarks}

We compare the algorithms on four standard benchmark functions for black-box optimization with dimensions ranging from 2 to 6. To obtain a more robust evaluation of the performance of each model, we generate multiple instances for each benchmark function by applying a random translation ($-0.1$--$0.1$), scaling ($0.9$--$1.1$), flipping, and dimension permutation in the input domain. 

The lef hand side of Figure~\ref{fig:benchmarks} shows the minimum observed function values achieved by the learned DNC optimizers, and contrasts these against the ones attained by Spearmint, TPE and SMAC. All methods appear to have similar performance with Spearming doing slightly better in low dimensions. As the dimension increases, we see that the DNC optimizers converge at at a much faster rate within the horizon of $T=100$ steps.

We also observe that DNC OI and DNC EI both outperform DNC with direct obsevations of the loss (DNC sum). It is encouraging that the curves for DNC OI and DNC EI are so close. While DNC EI is distilling a popular acquisition function from the EI literature, the DNC OI variant is much easier to train as it never requires the GP computations necessary to construct the EI acquisition function. 

The right hand side of Figure~\ref{fig:benchmarks} shows that
the neural network optimizers run about $10^4$ times faster than Spearmint and $10^2$ times faster than TPE and SMAC with the DNC architecture. There is an additional 5 times speedup when using the LSTM architecture, as shown in Table~\ref{tab:run_time}. The negligible runtime of our optimizers suggests new areas of application for global optimization methods that require both high sample efficiency and real-time performance.

\begin{table}[htbp]
\setlength{\tabcolsep}{4pt}
\centering
\caption{Run-time (seconds) for 100 iterations excluding the black-box function evaluation time.}
\label{tab:run_time}
\vspace{2mm}
\begin{tabular}{c|ccc|cc}
& Spearmint & TPE & SMAC & DNC & LSTM  \\
\hline
Branin & 1239 & 16.3 & 16.3 & 0.1 & 0.02 \\
Goldstein  & 1238 & 16.2 & 16.2 & 0.1 & 0.02 \\
Hartmann 3 & 1524 & 19.3 & 19.3 & 0.1 & 0.02 \\
Hartmann 6 & 2768 & 20.8 & 20.8 & 0.1 & 0.02 \\
\end{tabular}
\end{table}


\subsection{Transfer to a Simple Control Problem}

We also consider an application to a simple reinforcement learning task described by \cite{hoffman:2009b}. In this problem we simulate a physical system consisting of a number of repellers which affect the fall of particles through a 2D-space. The goal is to direct the path of the particles through high reward regions of the state space and maximize the accumulated discounted reward. The four-dimensional state-space in this problem consists of a particle's position and velocity. The path of the particles can be controlled by the placement of repellers which push the particles directly away with a force inversely proportional to their distance from the particle. At each time step the particle's position and velocity are updated using simple deterministic physical forward simulation. The control policy for this problem consists of 3 learned parameters for each repeller: 2d location and the strength of the repeller. 

\begin{figure}[h]%
\centering
\includegraphics[width=0.38\textwidth]{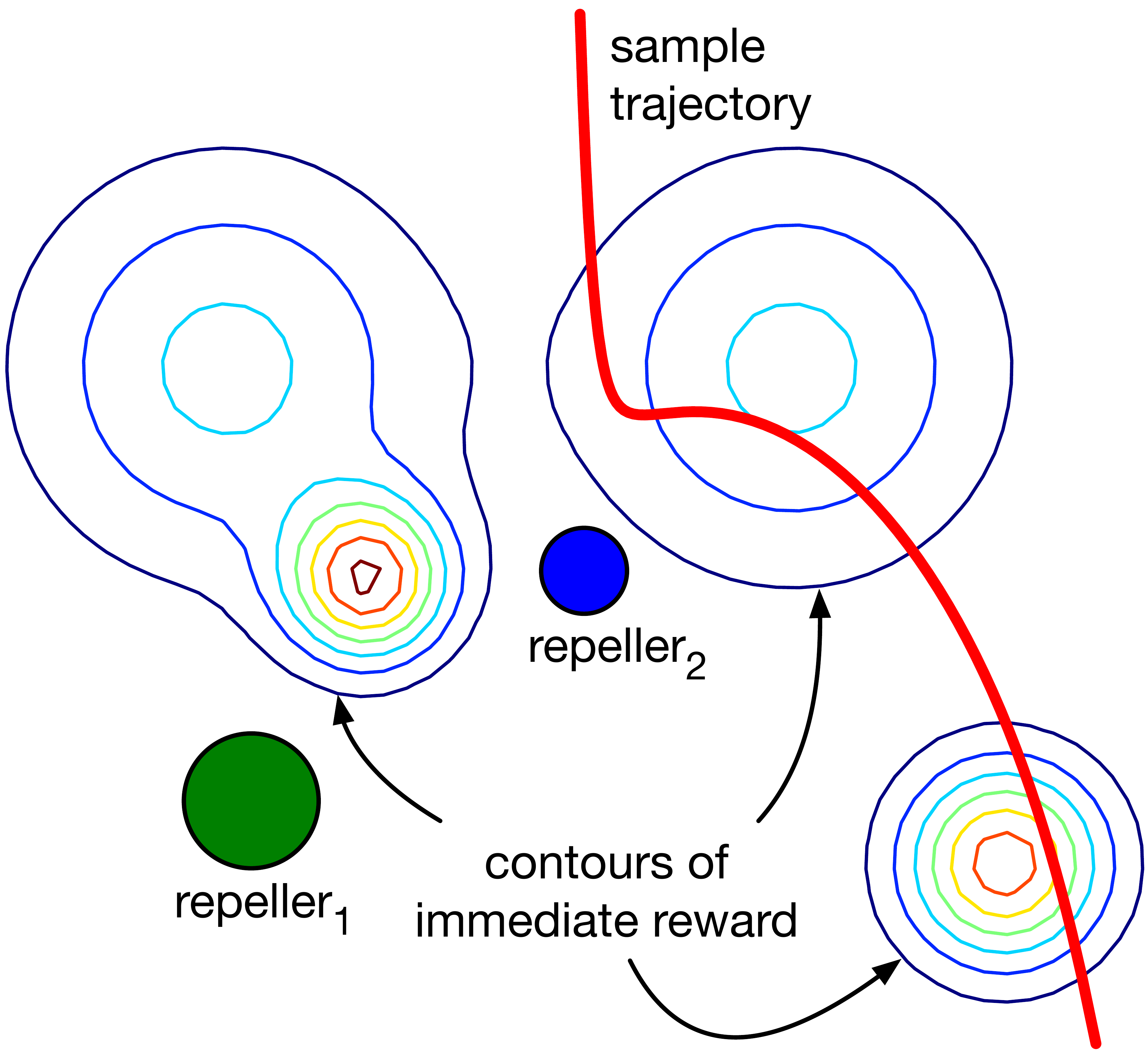} 
\\
\vspace{2mm}
\centering
\includegraphics[trim=40 0 300 0,clip,width=0.6\textwidth]{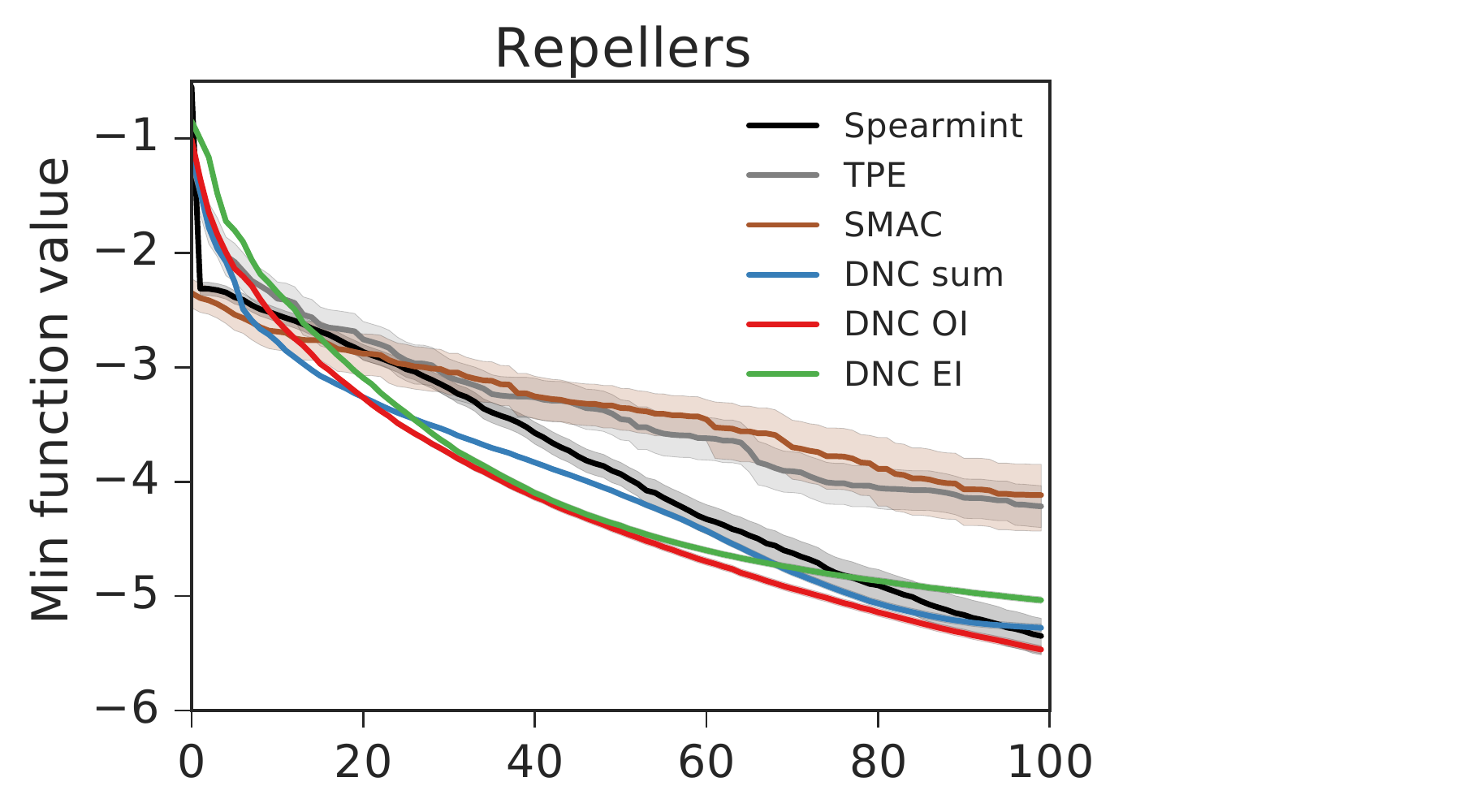}
\caption{Top: An example trajectory of a falling particle in red, where solid circles show the position and strength of the two repellers and contour lines show the reward function. The aim to to position and choose the strength of the repellers so that the particle spends more time in regions of high reward.
Bottom: The results of each method on optimizing the controller by direct policy search. Here, the learned DNC OI optimizer appears to have an edge over the other techniques.}
\label{fig:Repellers}
\end{figure}

In our experiments we consider a problem with 2 repellers, i.e.\ 6 parameters. An example trajectory along with the reward structure (contours) and repeller positions (circles) is displayed in Figure \ref{fig:Repellers}. We apply the same perturbation as in the previous subsection to study the average performance. The loss (minimal negative reward) of all models are also plotted in Figure \ref{fig:Repellers}. Neural network models outperform all the other competitors in this problem.

\begin{figure*}[htb]%
\centering
\includegraphics[width=\textwidth]{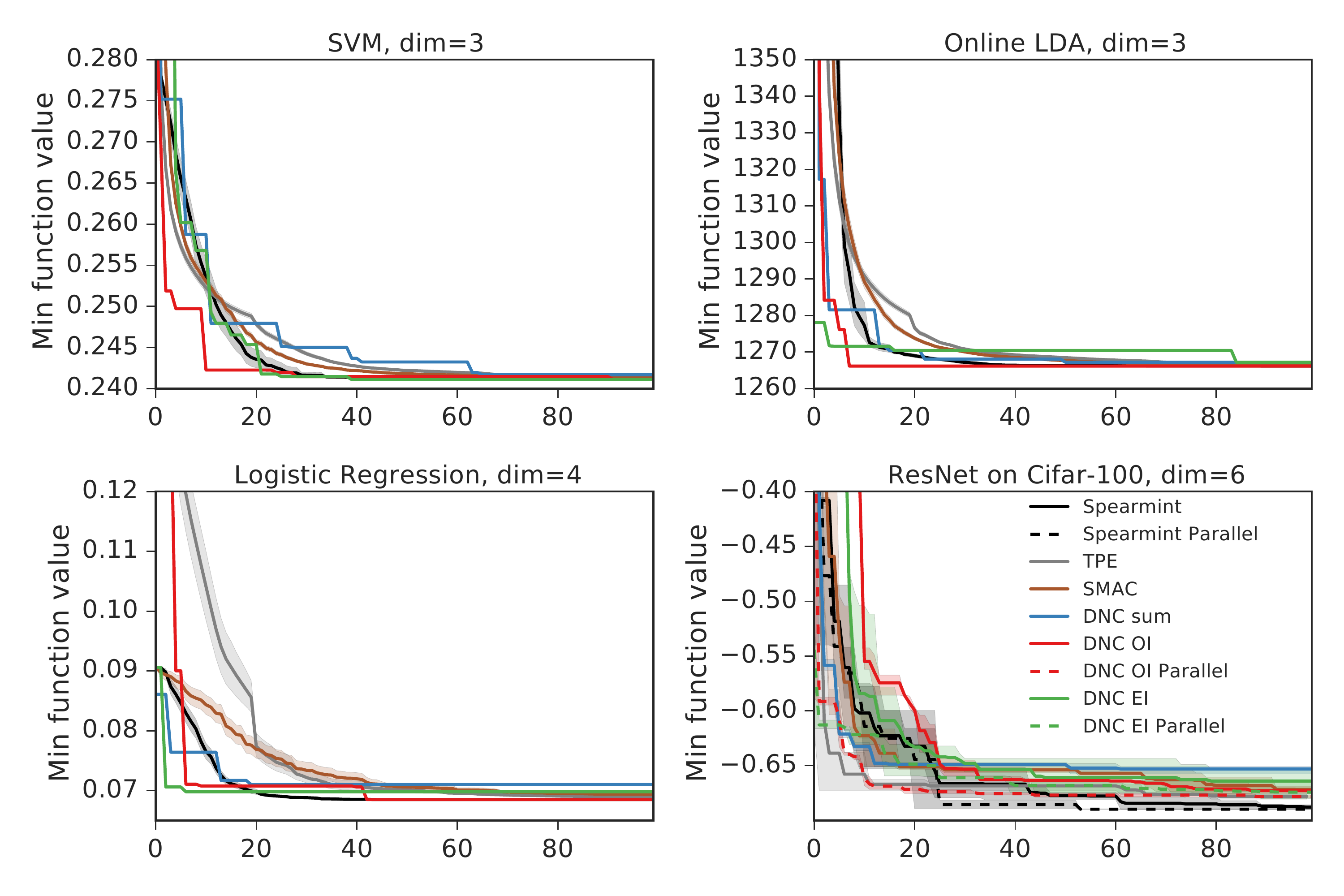}
\caption{Average test loss, with $95\%$ confidence intervals, for the SVM, online LDA, and logistic regression hyper-parameter tuning benchmarks. The bottom-right plot shows the performance of all methods on the problem of tuning a residual network, demonstrating that the learned DNC optimizers are close in performance to the engineered optimizers, and that the faster parallel versions work comparably well.}
\label{fig:ml_hyperparams}
\end{figure*}


\subsection{Transfer to ML Hyper-parameter Tuning}

Lastly, we consider hyper-parameter tuning for machine learning problems. We include the three standard benchmarks in the HPOLib package \citep{EggFeuBerSnoHooHutLey13}: SVM, online LDA, and logistic regression with 3, 3, and 4 hyper-parameters respectively. We also consider the problem of training a 6-hyper-parameter residual network for classification on the CIFAR-100 dataset. 

For the first three problems, the objective functions have already been pre-computed on a grid of hyper-parameter values, and therefore evaluation with different random seeds (100 for Spearmint, 1000 for TPE and SMAC) is cheap.  For the last experiment, however, it takes at least 16 GPU hours to evaluate one hyper-parameter setting. For this reason, we test the parallel proposal idea introduced in Section 2.3, with 5 parallel proposal mechanisms. This approach is about five times more efficient.

For the first three tasks, our model is run once because the setup is deterministic. For the residual network task, there is some random variation so we consider three runs per method.

The results are shown in Figure \ref{fig:ml_hyperparams}. The plots report the negative accuracy against number of function evaluations up to a horizon of $T=100$. The neural network models especially when trained with observed improvement show competitive performance against the engineered solutions. In the ResNet experiment, we also compare our sequential DNC optimizers with the parallel versions with 5 workers. In this experiment we find that the learned and engineered parallel optimizers perform as well if not slightly better than the sequential ones. These minor differences arise from random variation.

\section{Conclusions and Future Work}

The experiments have shown that up to the training horizon the learned RNN optimizers are able to match the performance of heavily engineered Bayesian optimization solutions, including Spearmint, SMAC and TPE. The trained RNNs rely on neither heuristics nor hyper-parameters when being deployed as black-box optimizers.  

The optimizers trained on synthetic functions were able to transfer successfully to a very wide class of black-box functions, associated with GP bandits, control, global optimization benchmarks, and hyper-parameter tuning.

The experiments have also shown that the RNNs are massively faster than other Bayesian optimization methods. Hence, for applications involving a known horizon and where speed is crucial, we recommend the use of the RNN optimizers. The parallel version of the algorithm also performed well when tuning the hyper-parameters of an expensive-to-train residual network.

However, the current RNN optimizers also have some shortcomings. 
Training for very long horizons is difficult. This issue was also documented recently in \cite{Duan2016}. We believe curriculum learning should be investigated as a way of overcoming this difficulty. In addition, a new model has to be trained for every input dimension with the current network architecture. While training optimizers for every dimension is not prohibitive in low dimensions, 
future works should extend the RNN structure to allow a variable input dimension. A promising solution is to serialize the input vectors along the search steps.

\setlength{\bibsep}{2.5pt}
{\small
\bibliographystyle{abbrvnat}
\bibliography{bayesopt,learning-to-learn,learning-to-experiment}
}


\end{document}